%% file: main.tex
\documentclass[10 pt, conference]{ieeeconf} 

\IEEEoverridecommandlockouts
\usepackage{cite}
\usepackage{subcaption}
\usepackage{graphicx}
\usepackage{amsmath,amssymb,amsfonts}
\usepackage{algorithmic}
\usepackage{graphicx}
\usepackage{textcomp}
\usepackage{xcolor}
\usepackage{verbatim}
\def\BibTeX{{\rm B\kern-.05em{\sc i\kern-.025em b}\kern-.08em
    T\kern-.1667em\lower.7ex\hbox{E}\kern-.125emX}}
\usepackage{listings}
\usepackage{url} 
\usepackage{hyperref}
\usepackage[british]{babel}
\usepackage{array}
\usepackage{multirow}
\usepackage{arydshln}
\usepackage[T1]{fontenc}
\usepackage{aecompl}
\usepackage{float}

\title{\LARGE \bf GRID: Scene-Graph-based Instruction-driven Robotic Task Planning}

\author{Zhe Ni$^{\dag}$, Xiaoxin Deng$^{\dag}$, Cong Tai$^{\dag}$, Xinyue Zhu, Qinghongbing Xie, Weihang Huang, Xiang Wu, Long Zeng$^{*}$
\\
\href{https://github.com/jackyzengl/GRID}{https://github.com/jackyzengl/GRID}%
    \thanks{$^{\dag}$Equal contribution.}
    \thanks{$^{*}$Corresponding author. (e-mail: zenglong@sz.tsinghua.edu.cn)}%
    \thanks{Zhe Ni, XiaoXin Deng, Cong Tai, Xinyue Zhu, Qinghongbing Xie, Weihang Huang, and Long Zeng are with Tsinghua Shenzhen International Graduate School, Tsinghua University, Shenzhen, China.}
    \thanks{Xiang Wu is with Shenzhen Pudu Technology Inc., Shenzhen, China.}
}

\begin{document}
\maketitle
\thispagestyle{empty}
\pagestyle{empty}




\input{abstract}
\input{introduction}

\input{relatedwork}

\input{problemstatement}
\input{method}

\input{dataset}

\input{experiments}
\input{conclusions}




\clearpage
\bibliographystyle{IEEEtran}
\bibliography{merged_nz_xxd_references_replace}
\end{document}

%% file: abstract.tex
\begin{abstract}


    Recent works have shown that Large Language Models (LLMs) can facilitate the grounding of instructions for robotic task planning.
    Despite this progress, most existing works have primarily focused on utilizing raw images to aid LLMs in understanding environmental information.
    However, this approach not only limits the scope of observation but also typically necessitates extensive multimodal data collection and large-scale models.
    In this paper, we propose a novel approach called \textbf{G}raph-based \textbf{R}obotic \textbf{I}nstruction \textbf{D}ecomposer (\textbf{GRID}), which leverages scene graphs instead of images to perceive global scene information and iteratively plan subtasks for a given instruction.
    Our method encodes object attributes and relationships in graphs through an LLM and Graph Attention Networks, integrating instruction features to predict subtasks consisting of pre-defined robot actions and target objects in the scene graph.
    This strategy enables robots to acquire semantic knowledge widely observed in the environment from the scene graph.
    To train and evaluate GRID, we establish a dataset construction pipeline to generate synthetic datasets for graph-based robotic task planning. 
    Experiments have shown that our method outperforms GPT-4 by over 25.4\% in subtask accuracy and 43.6\% in task accuracy.
    Moreover, our method achieves a real-time speed of 0.11s per inference.
    Experiments conducted on datasets of unseen scenes and scenes with varying numbers of objects demonstrate that the task accuracy of GRID declined by at most 3.8\%, showcasing its robust cross-scene generalization ability.
    We validate our method in both physical simulation and the real world. 
    More details can be found on the project page \href{https://jackyzengl.github.io/GRID.github.io/}{https://jackyzengl.github.io/GRID.github.io/}.

\end{abstract}

%% file: introduction.tex
\section{Introduction}

Recent advancements in large language models (LLMs) have shown promising outcomes in robotic task planning with instructions\cite{shridhar_cliport_2022, driess_palm-e_2023, brohan_rt-2_2023}.
LLM-based robotic task planning necessitates a cohesive comprehension of instructions and semantic knowledge of the environment. 
The primary challenge lies in adequately representing environmental information for LLM-based approaches to leverage, given that LLMs are designed for processing textual content.

In recent years, several studies have concentrated on integrating the visual modality to convey environmental knowledge. 
Some efforts combined the outcomes of vision models and LLMs in parallel \cite{ahn_as_2022, nair_r3m_2022, huang_voxposer_2023}, while others reasonably aligned the two modalities using vision-language models (VLMs) \cite{driess_palm-e_2023, brohan_rt-2_2023, jin_alphablock_2023, mees_grounding_2023, mu_embodiedgpt_2023}.
However, employing images as input limits the comprehension of global scenario information, and the feature alignment between pixels and instructions is challenging.

To overcome the limitations of images, a widely observed and rich-semantic representation of a scene, called the scene graph, has been introduced to robotics research \cite{armeni_3d_2019, rosinol_3d_2020}.
Recently, faster and more precise methods for scene graph generation \cite{wald_learning_2020, wu_scenegraphfusion_2021, wu2023incremental, koch2024open3dsg} facilitated the widespread utilization of graphs for long-term task planning \cite{zhu_hierarchical_2021, amiri_reasoning_2022, han_scene_2022}.
The applications of scene graphs \cite{kenghagho_robotvqa_2020, liu_bird_2023} have demonstrated that the introduction of graphs effectively addresses issues such as catastrophic forgetting, hallucination predictions, language ambiguity, and the lack of real-world environmental experience in LLM models.
However, graph-based task planning with natural language instructions has not garnered sufficient attention.

\begin{figure}[t]
    \centering
    \includegraphics[width=\columnwidth]{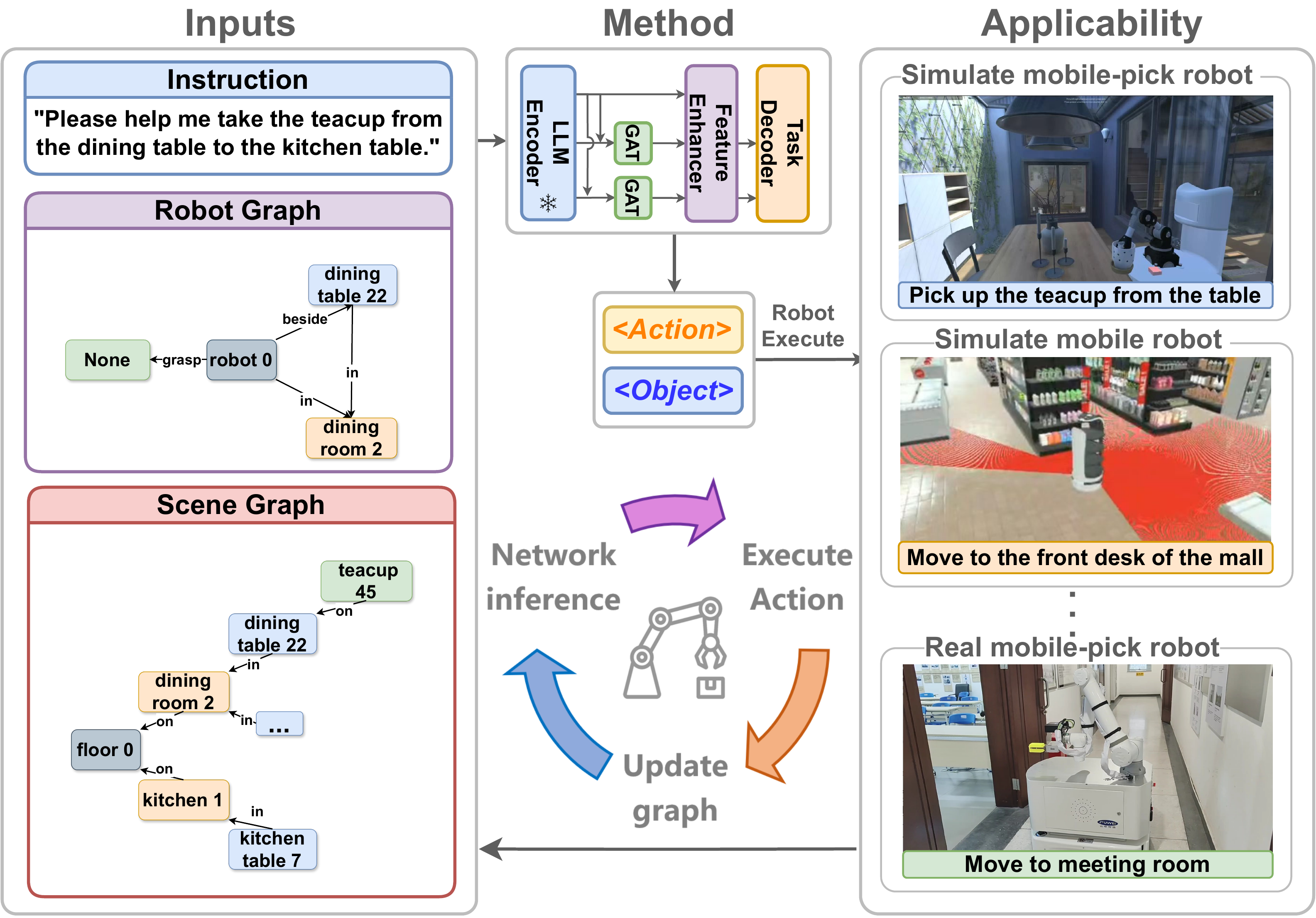}
    \caption{
    Our network, GRID, leverages instructions, scene graphs, and robot graphs as inputs for robotic task planning.
    Both environmental knowledge and the robot’s state are densely represented through graphs.
    The robot iteratively updates the graphs and executes the subtasks planned by GRID until completing the entire task.
    GRID can be deployed to robots in different forms,  operating effectively in various environments. 
    }
    \vspace{-10pt}
    \label{fig-First_Pic}
\end{figure}

Unlike previous works, we utilize scene graphs to represent the environment for robotic task planning.
In scene graphs, objects and their relationships within a scene are structured into graph nodes and edges.
To provide a precise portrayal of the robot's status, we separate the robot from the scene graph to create a robot graph, which contains the robot's location, nearby objects, and grasped objects.
We introduce a lightweight transformer-based model \textbf{GRID}, designed for deployment on offline embodied agents. 
As illustrated in Fig. \ref{fig-First_Pic}, the model takes instruction, scene graph, and robot graph as inputs, subsequently determining the subtask for the robot to execute.
In GRID, the instruction and both graphs are mapped into a unified latent space by a shared-weight LLM encoder named $INSTRUCTOR$\cite{su_one_2023}.
Then the encoded graph nodes and their relationships are refined by graph attention network (GAT) modules. 
Integrating the outputs from GAT and the encoded instruction with a cross-attention-based feature enhancer, the resultant enhanced features are fed into a transformer-based task decoder to yield the robot subtask.
Compared with the holistic sentence form, our subtask expressed as \texttt{<action>-<object>} pair form requires less enumeration to cover diverse object categories.
GRID iteratively plans the subtask, enabling it to respond to real-time scene changes and human interference, thereby correcting unfinished tasks.

Since existing scene graph datasets are not designed for robotic task planning \cite{mees_calvin_2022}, we develop a synthetic dataset construction pipeline to generate scene graph datasets for instruction-driven robotic task planning.
We use subtask accuracy and task accuracy as evaluation metrics. 
Subtask accuracy is the proportion of the subtasks that simultaneously predict the correct action and object.
Task accuracy refers to the proportion of instruction tasks for which all associated subtasks are predicted correctly and sequentially.
Evaluations on our datasets have shown that our method outperforms GPT-4 by over 25.4\% in subtask accuracy and 43.6\% in task accuracy.
The small parameter size of GRID enables real-time offline inference with a time consumption of 0.11s per inference.
To test the model's generalization ability, we constructed five datasets with scenes of different sizes and an unseen scenes dataset (i.e. all the scenes were unseen during the training phase).
The maximum decrease in task accuracy on these datasets was only 3.8\% without additional training.
Tests in simulated environments and the real world demonstrate that our approach enables cross-room task planning, which is challenging for visual-based approaches.


\textbf{Summary of Contributions:}
\begin{itemize}
    \item We first introduce scene graphs to facilitate instruction-driven robotic task planning by leveraging the graphs' ability to comprehend wide-perspective and rich semantic knowledge of the environment.
    \item We propose a novel GAT-based network called GRID, which takes instructions, robot graphs, and scene graphs as inputs.
    GRID outperforms GPT-4 by over 43.6\% in task accuracy while maintaining a low inference time cost of only 0.11s.
    \item We construct a synthetic dataset generation pipeline to produce datasets for scene-graph-based instruction-driven robotic task planning. 
\end{itemize}

%% file: relatedwork.tex
\section{Related Work}

\subsection{Scene Graph in Robotic Research}

Scene graphs are rich in semantic and dense representations of scenes\cite{bae_survey_2023}.
Previous works have explored robotic applications of scene graphs in navigation and task planning.
Sepulveda \emph{et al.}\cite{sepulveda_deep_2018} and Ravichandran \emph{et al.}\cite{ravichandran_hierarchical_2022} utilized scene graphs to enable platform-agnostic mobile robot navigation.
Jiao \emph{et al.}\cite{jiao_sequential_2022} demonstrated that scene graphs can assist robots in understanding the semantics of a scene to perform task planning.
Amiri \emph{et al.}\cite{amiri_reasoning_2022} and Han \emph{et al.}\cite{han_scene_2022} showed that robots can extract more contextual information from scene graphs than from a single image, benefiting from the wide observable domains of scene graphs, which leads to improved long-term task planning.
SayPlan \cite{rana_sayplan_2023} is closely related to our work.
They utilized an online LLM to process the collapsed scene graph and incorporated simulated feedback.
However, the text-based output lacks object IDs, making it unable to handle multiple objects of the same type, and the LLM's hallucination may impact executability.
In the scene graph research field, instruction-driven robotic task planning has not received sufficient attention.
To our knowledge, we are the first to use graph networks to process scene graphs and integrate LLMs to address instruction-driven robotic task planning.

\subsection{Understand Scene Information With LLMs}
Recent advancements in LLMs have promoted robotic systems' comprehension of instructions, where perceiving scene information is essential for solving a wide range of grounded real-world robotics problems \cite{tellex_robots_2020}. 
Previous studies have explored various approaches to incorporate the visual modality, aiming to obtain richer scene information, with the primary challenge lying in fusing vision and text modalities.
Ahn \emph{et al.}\cite{ahn_as_2022} and Huang \emph{et al.}\cite{huang_voxposer_2023} multiplied the outputs of visual models and LLMs to integrate the two.
Dorbala \emph{et al.}\cite{dorbala_can_2023} converted results from distinct visual modules (e.g., object detection, image captioning) into text modality prompts.
Inspired by CLIP\cite{radford_learning_2021}, several studies have aligned both modalities into the same space\cite{shridhar_cliport_2022, nair_r3m_2022, jang_bc-z_2022, lynch_interactive_2022, mees_grounding_2023}.
Recent research indicates that incorporating features from one modality into another within the model yields better performance. 
Brohan \emph{et al.}\cite{brohan_rt-1_2022} inserted instruction embeddings into a visual model, whereas Jin \emph{et al.}\cite{jin_alphablock_2023} and Driess \emph{et al.}\cite{driess_palm-e_2023} projected visual features onto a large-scale LLM to leverage its inference capabilities.
However, these models extract sparse and low-level semantic knowledge from raw images, rely heavily on large-scale data for training, and face challenges in analyzing information beyond what is depicted in the images.






\vspace{10pt}

In contrast, we utilize scene graphs instead of images in instruction-driven robotic task planning. 
Our approach enables the direct utilization of LLM to process rich semantic knowledge within scene graphs, eliminating the need for additional vision-text alignment or massive multi-modal data.
Additionally, scene graphs offer a broader observable domain to embodied agents compared to images, enabling robots to navigate to various locations and perform tasks beyond the immediate field of view.

%% file: problemstatement.tex
\section{Problem Statement}

\begin{figure*}[t]
    \centering
    \includegraphics[width=\textwidth]{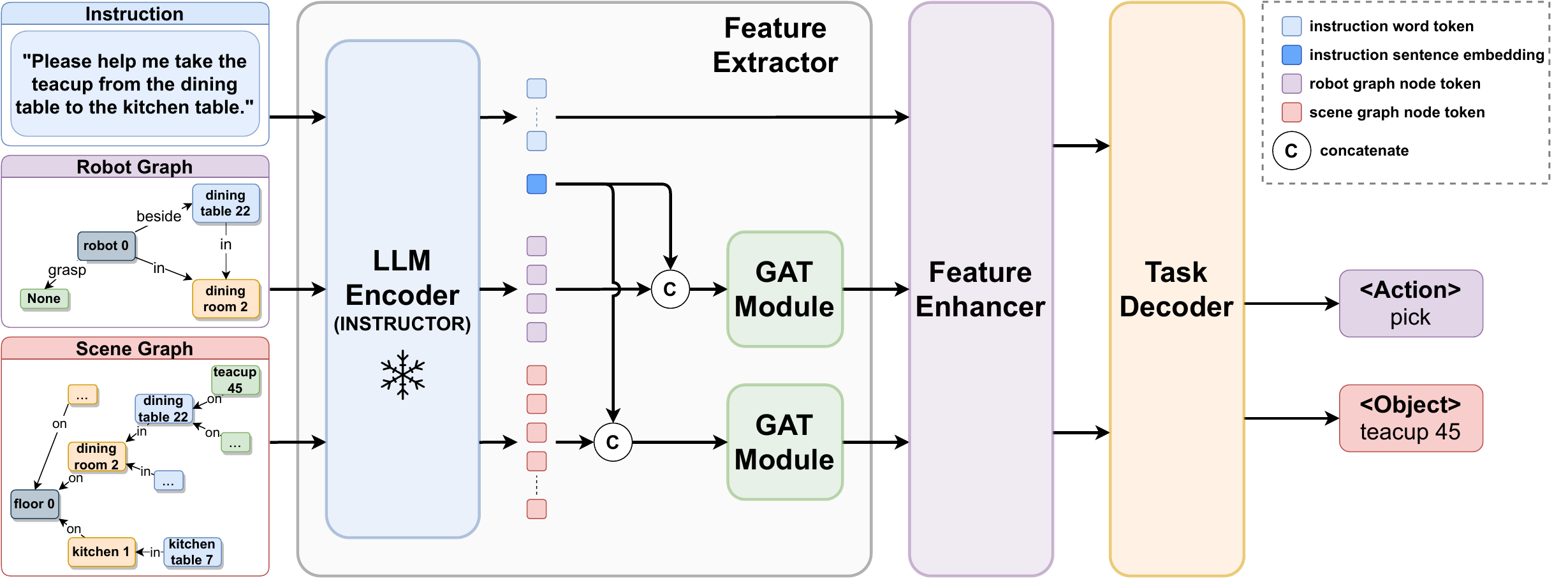}
    \caption{
        The architecture diagram of GRID. 
        The instruction, robot graph, and scene graph are all transformed into tokens through $INSTRUCTOR$\cite{su_one_2023}. 
        Subsequently, GAT modules extract structural information from the graphs. 
        The resulting tokens undergo reinforcement by a feature enhancer and are then fed into a task decoder, ultimately generating outputs for action and object in ID form.
    }
    \vspace{-10pt}
    \label{fig-Model_Overview}
\end{figure*}

We frame the graph-based instruction-driven task planning problem as follows: given an instruction, a robot graph, and a scene graph, our objective is to predict the subtask required for the robot to execute in the current stage. 
Typically, the instruction is a human-language task that can be decomposed into a series of subtasks.  

The scene graph $\mathbf{s}$ and robot graph $\mathbf{r}$ comprise multiple nodes $\mathbf{n}$ and edges $\mathbf{e}$. 
Nodes denote the objects within the environment characterized by attributes such as colors, positions, and other properties.
Edges depict the relationships between objects. 
They are formulated as:
\vspace{-2pt}
\begin{equation}
    \begin{aligned}
        \mathbf{s} &= \langle \mathbf{n}^S, \mathbf{e}^S \rangle ,\\
        \mathbf{r} &= \langle \mathbf{n}^R, \mathbf{e}^R \rangle , 
    \end{aligned}
\end{equation}
where $\mathbf{n}^S, \mathbf{e}^S$ represent the nodes and edges of the scene graph respectively, while $\mathbf{n}^R, \mathbf{e}^R$ denote the nodes and edges of the robot graph.

Let $\mathbb{I}$ denote the space of instructions, $\mathbb{A}$ be the action space of robot skills, and $\mathbb{O} = \left[{0, M} \right)$ represent the set of IDs of all nodes in the scene graph, where $M$ is the number of nodes in the scene graph. 
The output subtask space $\mathbb{P} = \mathbb{A}\times \mathbb{O}$, where $\times$ denotes the Cartesian product.

At stage $t$, 
given $\mathbf{r}^t, \mathbf{s}^t$ and a human-language instruction $I \in \mathbb{I}$, let $\hat{\pi}^t \in \mathbb{P}$ represent the predicted subtask, $\hat{a}^t \in \mathbb{A}$ denote the predicted action, and $\hat{o}^t \in \mathbb{O}$ be the ID of the predicted node in $\mathbf{s}^t$.
The objective function is formulated as follows:
\vspace{-2pt}
\begin{equation}\label{eqn-objective-pi}
    I, \mathbf{r}^t, \mathbf{s}^t \xrightarrow[]{infer\ subtask} \hat{\pi}^t = \langle \hat{a}^t, \hat{o}^t \rangle 
\end{equation}

After the robot interacts with the environment by executing the subtask predicted at stage $t$, the graphs are updated to the next stage $\mathbf{r}^{t+1}, \mathbf{s}^{t+1}$. 
The robot graph can be updated by the robot sensor algorithm, while the scene graph can be constructed and updated using scene graph generation methods \cite{armeni_3d_2019, rosinol_3d_2020, wald_learning_2020, wu_scenegraphfusion_2021, wu2023incremental, koch2024open3dsg}.
\vspace{-2pt}
\begin{equation}
    \begin{aligned}
        \mathbf{r}^{t} \xrightarrow[]{update}  \mathbf{r}^{t+1}\\
        \mathbf{s}^{t} \xrightarrow[]{update}  \mathbf{s}^{t+1}
    \end{aligned}
\end{equation}

The prediction iterates until all subtasks associated with the given instruction $I$ have been executed.


%% file: method.tex
\section{Method}

An overview of our method is depicted in Fig. \ref{fig-Model_Overview}. 
Our model $\mathbf{GRID}$ predicts a pair of action and object ID $\langle \hat{a}^{t}, \hat{o}^{t}\rangle$ for a given instruction, robot graph, and scene graph triad $\langle I, \mathbf{r} ^{t}, \mathbf{s} ^{t} \rangle$ at stage $t$:
\vspace{-2pt}
\begin{equation}
    \langle \hat{a}^{t} , \hat{o}^{t}\rangle
    = \mathbf{GRID}(\langle I, \mathbf{r} ^{t}, \mathbf{s} ^{t} \rangle)
\end{equation}
For clarity, the stage label $t$ will be omitted in the formulas in the subsequent discussions.

GRID employs an encoder-decoder architecture. 
The feature extractor comprises a shared-weight LLM and two GAT modules. 
The instruction, robot graph, and scene graph all undergo processing by the LLM, serving as a text backbone for extracting raw semantic features. 
Two GAT modules are utilized to extract structural features from the respective graphs (Sec. \ref{Sec-Feature_Extraction}). 
The feature enhancer collectively reinforces the overlapping information of instructions and graphs (Sec. \ref{Sec-Feature_Enhancer}).
The task decoder comprehensively analyzes the features and generates the predicted results (Sec. \ref{Sec-Task_Decoder}).

\subsection{Feature Extractor}\label{Sec-Feature_Extraction}

We employ a frozen pre-trained LLM named $INSTRUCTOR$ as a shared-weight text backbone. 
Given a sentence of length $m$, it generates task-aware embeddings\cite{su_one_2023}. 
These embeddings consist of a sequence of word tokens $ \mathbf{w} = (w_1, ..., w_m)$ and a sentence embedding $y$, which represents the features of the entire sentence. 
Employing the same text backbone enables the features of instructions and graphs to be mapped into the same latent space without requiring additional alignment. 
For instructions, we compute both word tokens $\mathbf{w}^{I} $ and sentence embedding $ y^{I} $:
\vspace{-2pt}
\begin{equation}
    \langle \mathbf{w}^{I}, y^{I} \rangle = INSTRUCTOR(I)
\end{equation}

For both robot graphs and scene graphs, each node's attributes are treated as distinct sentences. 
Only the sentence embedding for each node is considered, which is interpreted as a node token:
\vspace{-2pt}
\begin{equation}
    \begin{aligned}
        & \mathbf{nt}^{R} = INSTRUCTOR(\mathbf{n}^{R}) ,\\
        & \mathbf{nt}^{S} = INSTRUCTOR(\mathbf{n}^{S}) ,
    \end{aligned}
\end{equation}
where $\mathbf{nt}^{R}$
denotes the sequence of node tokens from nodes in the robot graph, and 
$\mathbf{nt}^{S}$
denotes the sequence of node tokens from nodes in the scene graph.

To incorporate instruction information when extracting graph features, we concatenate the sentence embedding of the instruction with every node token.
Then, the node token, along with the edge information in each graph, is inputted to their respective graph attention network (GAT) modules \cite{velickovic_graph_2018}:
\vspace{-2pt}
\begin{equation}
    \begin{aligned}
        & \mathbf{y}^{R} = GAT_{robot}(concat(\mathbf{nt}^{R} , y^{I}), \mathbf{e}^{R}) ,\\
        & \mathbf{y}^{S} = GAT_{scene}(concat(\mathbf{nt}^{S} , y^{I}), \mathbf{e}^{S}) ,
    \end{aligned}
\end{equation} 
where $GAT_{robot}, GAT_{scene}$ denote GAT modules for the robot and scene graph respectively.


\subsection{Feature Enhancer}\label{Sec-Feature_Enhancer}

\begin{figure}[t]
    \centering
    \includegraphics[width=0.8\columnwidth]{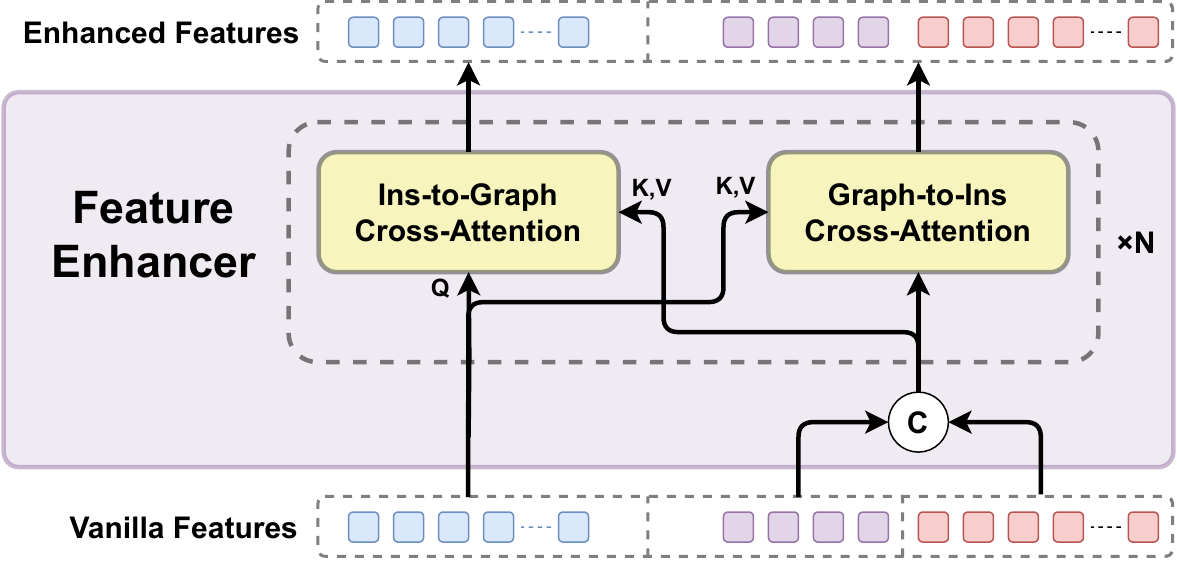}
    \caption{
        Vanilla instruction and graph features are fed into N layers of parallel cross-attention to enhance crossover information between instructions and graphs.
    }
    \vspace{-10pt}
    \label{fig-Feature_Enhancer}
\end{figure}

Inspired by Grounding DINO\cite{liu_grounding_2023}, we have designed a feature enhancer, as illustrated in Fig. \ref{fig-Feature_Enhancer}. 
The feature enhancer utilizes a stack of parallel multi-head cross-attention layers $CA(Q; K; V)$.


$ CA_{I}, CA_{g} $ denote instruction-to-graph and graph-to-instruction multi-head cross-attention layers with residual connections respectively. 
At layer $l$, $CA_{I}$ takes the word tokens from instructions $\mathbf{w}^{I}_{l-1}$ as queries, and the node tokens from graphs $\langle \mathbf{y}^{R}_{l-1}, \mathbf{y}^{S}_{l-1} \rangle$ as keys and values. 
Conversely, in $CA_{g}$, $\mathbf{w}^{I}_{l-1}$ serve as keys and values, while $\langle \mathbf{y}^{R}_{l-1}, \mathbf{y}^{S}_{l-1} \rangle$ functions as queries:
\vspace{-2pt}
\begin{equation}
    \begin{aligned}
        \mathbf{w}^{I}_{l} = CA_{I}(\mathbf{w}^{I}_{l-1}; \langle \mathbf{y}^{R}_{l-1}, \mathbf{y}^{S}_{l-1} \rangle; \langle \mathbf{y}^{R}_{l-1}, \mathbf{y}^{S}_{l-1} \rangle )  \\
        \langle \mathbf{y}^{R}_{l}, \mathbf{y}^{S}_{l} \rangle = CA_{g}(\langle \mathbf{y}^{R}_{l-1}, \mathbf{y}^{S}_{l-1} \rangle; \mathbf{w}^{I}_{l-1}; \mathbf{w}^{I}_{l-1})
    \end{aligned}
\end{equation}

These structures emphasize the crossover information between instructions and graphs, particularly when the same object appears in both. 
After traversing the final layer of the feature enhancer, the tokens from the instruction act as fusion features $\mathbf{f}^I = \mathbf{w}^{I}_{last}$, and the tokens from graphs serve as graph queries $\langle \mathbf{q}^{R}, \mathbf{q}^{S} \rangle = \langle \mathbf{y}^{R}_{last}, \mathbf{y}^{S}_{last} \rangle$.
Subsequently, they are inputted into the task decoder.

\subsection{Task Decoder}\label{Sec-Task_Decoder}

\begin{figure}[t]
    \centering
    \includegraphics[width=0.8\columnwidth]{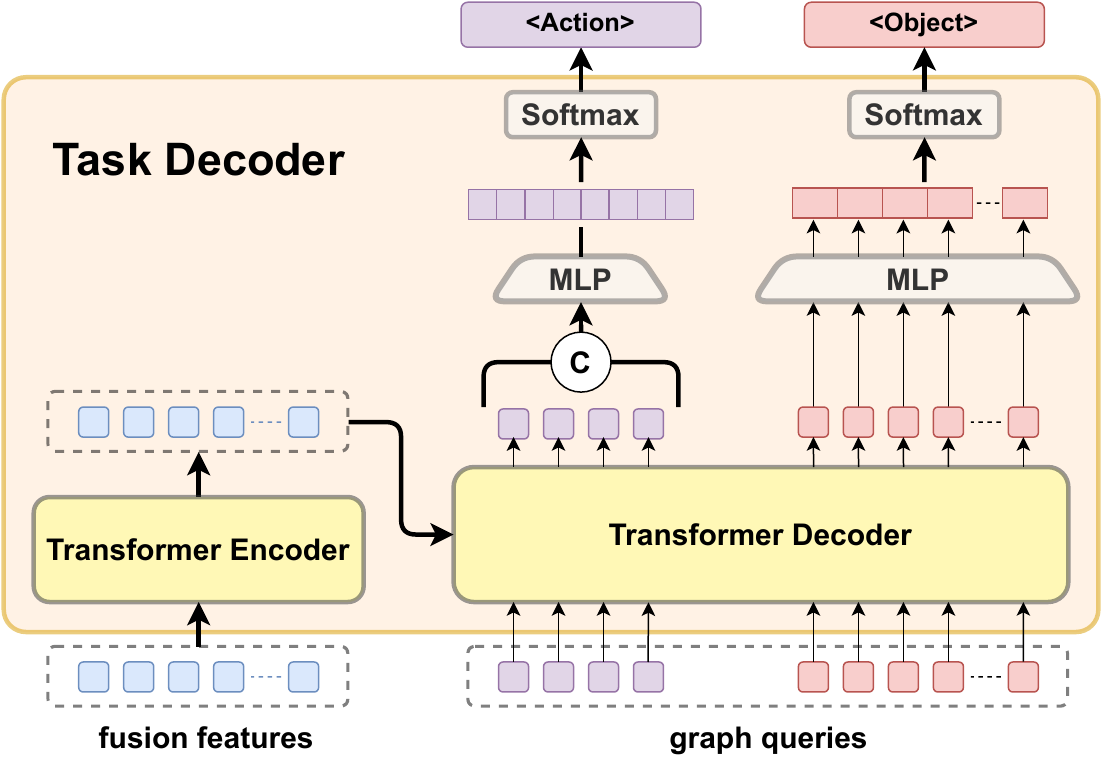}
    \caption{
        The enhanced tokens are segregated into fusion features and graph queries, which are input into the transformer encoder and decoder, respectively. 
        The tokens from the robot graph are mapped to scores for each action, and each token from the scene graph is converted to the score for the corresponding node.
    }
    \vspace{-10pt}
    \label{fig-Task_Decoder}
\end{figure}

We develop a task decoder to integrate instruction and graph features for predicting the action and object of the subtask, as depicted in Fig. \ref{fig-Task_Decoder}. 

Different from treating the subtask as an integral sentence in previous works, we divide it into \texttt{<action>} and \texttt{<object>} as two parts. 
This division enables diverse combinations of robot "atomic" skills with objects in the scene, without being limited by the types of objects like SayCan\cite{ahn_as_2022} (which covers 551 subtasks spanning only 17 objects) or requiring enumeration of a large number of subtasks like in LAVA\cite{lynch_interactive_2022} (87K subtasks). 
We use IDs from the scene graph to uniquely identify objects instead of relying on textual descriptions, thus avoiding ambiguities for downstream modules\cite{driess_palm-e_2023, chalvatzaki_learning_2023, dorbala_can_2023}.

Fusion features are fed into a transformer encoder $Enc(src)$. 
Graph queries are fed into a transformer decoder $Dec(tgt, src)$. The output embeddings enter the action branch and object branch respectively: 
\begin{equation}
    \langle \mathbf{f}^{R}, \mathbf{f}^{S} \rangle = Dec(\langle \mathbf{q}^{R}, \mathbf{q}^{S} \rangle, Enc(\mathbf{f}^I)),
\end{equation}
where $\mathbf{f}^{R}$ denotes the output embeddings of the transformer decoder from queries $\mathbf{q}^{R}$, and $\mathbf{f}^{S}$ from queries $\mathbf{q}^{S}$.

Given the close relationship between the current state of the robot and the subsequent action, the output embeddings from the robot graph $\mathbf{f}^{S}$ are fed into an MLP (Multi-Layer Perceptron) after concatenation to yield $N_{act}$ scores. 
The action associated with the highest score will be selected as the action to be executed: 
\vspace{-2pt}
\begin{equation}
    \hat{a} = softmax(MLP_{act}(concat(\mathbf{f}^{R}))) 
\end{equation}

The output embeddings $\mathbf{f}^{S}$ from each node in the scene graph undergo individual processing through an MLP to derive node-specific scores.
These scores indicate the probabilities of each object becoming the operated object:
\vspace{-2pt}
\begin{equation}
    \hat{o} = softmax(MLP_{obj}(\mathbf{f}^{S}))
\end{equation}

\begin{table}[t]
    \centering
    \begin{tabular}{ll}
        \hline
        \textbf{Subtask Type}      & \textbf{Count} \\ \hline
        \texttt{<move> - <object>}                 & 18347 \\
        \texttt{<pick> - <object>}             & 11811 \\
        \texttt{<place to> - <object>}             & 17924 \\
        \texttt{<revolute open> - <object>}        & 3456  \\
        \texttt{<revolute close> - <object>}       & 3456  \\
        \texttt{<longitudinal open> - <object>}    & 3503  \\
        \texttt{<longitudinal close> - <object>}   & 3503  \\
        \texttt{<finish> - <object>}               & 19981 \\ \hline
        \textbf{Total}                & \textbf{81981} \\ \hline
    \end{tabular}
    \caption{The list of subtask types in the 70-object dataset, along with their respective counts.
    Subtasks are classified into different types based on their actions, as each is associated with a unique action.
    }
    \vspace{-10pt}
    \label{tab-Subtask_Action_Count}
\end{table}

\begin{table*}[t]
    \centering
    \begin{tabular}{|p{0.95\textwidth}|}
        \hline
        \#Role: You are an instruction planning model for a mobile grasping robot, you aim to break down high-level instruction into subtasks for the robot to execute. 
        \\\#Output Restriction:  Your output is in a form of \textless action\textgreater \textless operating object name\textgreater \textless operating object id\textgreater , where \textless action\textgreater is one of the actions in 
        ['move', 'pick', ...], 
        The meaning of the actions: move: controls the robot's movement. pick: ...\quad The \textless operating object name\textgreater  \textless operating object id\textgreater are the object and ids given in the prompt \#Scene ...
        \\\#Scene: The current scene has objects with ids: \textless \texttt{scene graph nodes description}\textgreater. 
        The relationships between objects in the scenes are: \textless \texttt{scene graph edges description}\textgreater 
        \\\#Robot: ...\quad \#Instruction: \textless \texttt{instruction}\textgreater 
        \\\#Task: Please tell me what the current subtask the robot should perform according to the scene and robot status. Write down your thought process.        
        \\\#Example1:
        \\\quad input: 
            \\\quad\quad\#Scene: The current scene has objects with IDs: floor 0, dining room 1, black dining table 2, purple display shelves 3, ... \quad The relationships between objects in the scenes are: dining room 1 is on floor 0, ...
            \\\quad\quad\#Robot: The objects and object IDs around the robot: robot 0, brown pen 2. \quad The relationships between objects around the robot are: pen 2 is grasped by robot 0.
            \\\quad\quad\#Instruction:
            Your goal is to get the object to the purple display shelves.
            \\\quad\quad\#Task: Please tell me ...
        \\\quad output:
            \\\quad\quad\#Think: Because I do not know where I am, so I should first move to purple display shelves... So output: move purple display shelves 3.
        \\\#Example2:
        ...
        \\
        \hline
    \end{tabular}
    \caption{A summarized version of the input prompt settings example of LLM-As-Planner. For the complete and detailed prompt settings example, please refer to the project page \href{https://jackyzengl.github.io/GRID.github.io/}{https://jackyzengl.github.io/GRID.github.io/}.}
    \vspace{-5pt}
    \label{tab-GPT_Prompt}
\end{table*}

\subsection{Loss Function}\label{Sec-Loss_Function}

We utilize the standard categorical cross-entropy loss for both the action branch $\mathcal{L}_{act}$ and the object branch $\mathcal{L}_{obj}$. 
The total loss function employed to train the complete model is defined as a weighted combination of these two losses with $L_1$ and $L_2$ regularization terms:
\vspace{-2pt}
\begin{equation}
\mathcal{L} = \alpha \mathcal{L}_{act} + \beta \mathcal{L}_{obj} + \gamma \sum_i |w_i| + \frac{\delta}{2} \sum_i w^2_i ,
\end{equation}

where $\mathcal{L}_{act} = \mathbf{CE} (\hat{a}^t , a^t_{gt}), \mathcal{L}_{obj} = \mathbf{CE} (\hat{o}^t , o^t_{gt})$, with the ground-truth action id $a^t_{gt}$ and object id $o^t_{gt}$. 
$\alpha, \beta$ denote loss weights, $\gamma, \delta$ denote $L_1$ and $L_2$ regularization strengths respectively, and $w_i$ denotes trainable weights in the network.

%% file: dataset.tex
\section{Dataset}\label{Sec-Dataset}
\begin{table*}[t]
    \center
    \begin{tabular}{cccccccc}
        \hline
        \multirow{2}{*}{\textbf{Model}} & \multirow{2}{*}{\textbf{\#Params}} & \textbf{30-object}           & \textbf{40-object}           & \textbf{50-object}           & \textbf{60-object}           & \textbf{70-object}           & \multirow{2}{*}{\textbf{\begin{tabular}[c]{@{}c@{}}Avg. \\ Infer Time\end{tabular}}} \\ \cline{3-7}
                                        &                                    & \textbf{S. A./T. A.} & \textbf{S. A./T. A.} & \textbf{S. A./T. A.} & \textbf{S. A./T. A.} & \textbf{S. A./T. A.} &                                           \\ \hline
        {LLM-As-Planner(LLaMA2)}                          & {{7B}}                                 & {{21.0\%/1.5\%}}                             &{{24.8\%/0.0\%}}                              &{{20.5\%/0.0\%}}                               &{{22.5\%/0.0\%}}                               &{{22.0\%/0.0\%}}                               &3.10s(Offline)                                        \\
        LLM-As-Planner(LLaMA2) & {70B} &{31.0\%/3.0\%} &{31.0\%/4.3\%} &{27.5\%/4.3\%} &{28.5\%/2.2\%} &{27.5\%/0.0\%} &4.87s (Online)\\
        LLM-As-Planner(GPT-4)                           & >1T                                & {55.3\%/16.7\%}       & {57.7\%/14.3\%}       & {54.5\%/11.1\%}       & {49.6\%/10.0\%}       & {47.5\%/8.0\%}        & 2.52s (Online)                               \\
        \textbf{GRID}                   & \textbf{2.3M+1.5B}                 & \textbf{81.6\%/60.3\%}       & \textbf{83.1\%/63.2\%}       & \textbf{82.8\%/63.0\%}       & \textbf{82.2\%/62.1\%}       & \textbf{83.0\%/64.1\%}       & \textbf{0.11s (Offline)}                  \\ \hline
    \end{tabular}
    \caption{Comparison between GRID and LLM-As-Planner on datasets with different numbers of objects in each scene. The offline inferences are conducted on one NVIDIA RTX4090 GPU. \textbf{S. A.} denotes subtask accuracy, \textbf{T. A.} denotes task accuracy. \textbf{Bold} values indicate the optimal data.}
    \vspace{-10pt}
    \label{tab-Less_Obj}
\end{table*}

\begin{table*}[t]
    \center
    \begin{tabular}{cccccc}
        \hline
        \textbf{Model}                            & \textbf{\#Params}  & \textbf{Act. Acc.} & \textbf{Obj. Acc.} & \textbf{Sub. Ac.} & \textbf{Task Acc.} \\ \hline
        \textbf{GRID}                            & \textbf{2.3M+1.5B} & \textbf{84.8}            & \textbf{94.1}            & \textbf{83.0}             & \textbf{64.1}          \\
        GRID w/o Feature Enhancer                & 2.2M+1.5B          & 83.2                     & 93.1                     & 81.0                      & 56.2                   \\
        GRID w/o GAT modules                     & 314K+1.5B          & 83.1                     & 93.3                     & 80.2                      & 55.5                   \\
        GRID w/o GAT modules \& Feature Enhancer & 245K+1.5B          & 76.2                     & 92.0                     & 73.0                      & 28.4                   \\ \hline
    \end{tabular}
    \caption{Ablation studies of key modules. 
            The number of parameters is recorded as the sum of trainable parameters and the parameters in the frozen LLM ($INSTRUCTOR$\cite{su_one_2023}).
            }
    \vspace{-10pt}
    \label{tab-Ablation}
\end{table*}

Graph-based robotic task planning takes an instruction, robot graph, and scene graph as inputs and plans the subtask in the current state as output. 
Similar datasets (ALFRED \cite{shridhar_alfred_2020}, CALVIN \cite{mees_calvin_2022}, ARNOLD\cite{gong_arnold_2023}, etc.) provide instructions and corresponding subtasks.
However, these datasets lack scene graphs and robot graphs corresponding to the subtasks or metrics indicating the completion of the entire task.
Thus, we developed a synthetic dataset construction pipeline to generate datasets consisting of four components: instructions, subtasks, robot graphs, and scene graphs. 
The subtasks in the dataset serve as the ground truth and are used for network training and evaluation. 
We pioneered the construction of datasets that establish the interconnections between instructions, robot graphs, scene graphs, and subtasks providing fundamental benchmarks for task planning in this domain.

To generate one set of data, we use a randomly generated domestic environment to initialize a scene graph, $\mathbf{s}_0$, as a blueprint. 
Meanwhile, the robot graph, $\mathbf{r}_0$, is initialized with a predefined robot state.
We then generate a list of feasible subtasks in this scene, e.g., pick-teacup.  
For each set of subtasks, corresponding human-language instructions are synthesized with the assistance of ChatGPT-4 in diverse expressions. 
These subtasks are used to iteratively update the blueprint $\mathbf{r}_0, \mathbf{s}_0$, obtaining the robot graphs and scene graphs for different stages.

We generated five datasets containing different object counts in each scene ranging from 30 to 70.
The dataset with 70 objects in each scene (70-object dataset) includes about 20K instruction tasks. 
These tasks can be decomposed into 82K subtasks, as shown in Tab. \ref{tab-Subtask_Action_Count}.
Additionally, we constructed an unseen scene dataset in which all scenes are not seen during the training phase.

%% file: experiments.tex
\section{Experiments}

In this section, we first describe the training details.
Then we compare our model with GPT-4 and validate the effectiveness of two key modules through ablation studies.
We also explore the generalization ability of the model with different datasets.
Finally, we demonstrate our approach both in the simulation environment and the real world.


\subsection{Training Details}\label{Sec-Training_Details}

GRID is implemented on the PyTorch toolbox. 
It is optimized using AdamW, with a peak learning rate of 1e-4 that decays according to a one-cycle learning rate schedule with a diversity factor of 10 and a final diversity factor of 1e-4.
We utilize 80\% of the data from the 70-object dataset, which is the largest dataset in terms of scale, for training, while the remaining 20\%, along with all other datasets, is used for evaluation. 
Our models are trained for 500 iterations with a total batch size of 240 on two NVIDIA RTX4090 GPUs. 
The loss weights are set to $\alpha = 5$ and $\beta = 25$, while the $L_1$ and $L_2$ regularization strengths are $\gamma = 0.2$ and $\delta = 0.8$, respectively.

\subsection{Evaluation}

\textbf{Metrics.}
To evaluate the performance of our method, we proposed four main accuracy metrics: action accuracy (Act. Acc.), object accuracy (Obj. Acc.), subtask accuracy (Sub. Acc.), and task accuracy (Task Acc.).
Considering each subtask sample as an independent prediction, action accuracy, and object accuracy measure the model's precision in predicting actions and objects, respectively.
Subtask accuracy records the probability of correctly predicting both action and object, while task accuracy represents the proportion of all subtasks being predicted sequentially and correctly.

\textbf{Comparison with baseline.}
Following Chalvatzaki \emph{et al.} \cite{chalvatzaki_learning_2023} and referring to SayPlan \cite{rana_sayplan_2023}, we evaluate the task planning performance of LLaMA2 and GPT-4 on our datasets.
The baseline methods settings are similar to LLM-As-Planner in SayPlan \cite{rana_sayplan_2023}, where we utilize LLMs to generate the sequence of plans.
Given that GPT-4 is one of the best-performing LLMs, its performance will surpass that of smaller-weight models like LLaMA2, Falcon, Vicuna, etc., making it a representative baseline.
The prompt settings of LLM-As-Planner are shown in Tab. \ref{tab-GPT_Prompt}.
We use the chain of thought (CoT) and examples to enhance performance because detailed prompting can elicit better results in LLMs.
To focus on the task planning ability without being influenced by the accuracy of scene graph generation, the input graphs are set to be complete and accurate.
The results, as shown in Tab. \ref{tab-Less_Obj}, demonstrate that in our 70-object dataset, our approach achieved 83.0\% subtask accuracy and 64.1\% task accuracy significantly outperforming GPT-4's 47.5\% and 8.0\% respectively.
This may be because our model leverages structural information in the graphs using GAT. 
Our model fuses and enhances the features of instructions and graphs repeatedly, allowing the model to focus on the objects mentioned in the instructions.
Meanwhile, predicting actions and objects separately can effectively reduce the difficulty of predicting the entire subtask.
Although GRID outperforms GPT-4 due to the benefits from trainable weights, it has significantly fewer parameters (2.3M+1.5B) and a shorter average inference time (0.11s) compared to GPT-4 (>1T, 2.52s) and LLaMA2 (7B, 3.10s; 70B4.87s). 
This enables its offline deployment and real-time inference on edge devices such as robots.

\textbf{Ablation studies.}
We perform ablation studies on the 70-object dataset to analyze the importance of the GAT modules (together with concatenation) and the feature enhancer.
The results shown in Tab. \ref{tab-Ablation} indicate that both the GAT modules and feature enhancer improve accuracy.
The accuracy of the complete model is significantly improved compared to the simplified model.
This may be because the full model both considers the relationships between objects in the graphs and sufficiently fuses the information from the instruction and graphs.

\textbf{Generalization to scenes of different sizes.}
The results of evaluation on five datasets with different numbers of objects in each scene in Tab. \ref{tab-Less_Obj} show that subtask and task accuracy of GRID are only slightly affected by the number of objects in each scene.
The subtask and task accuracy of GPT-4 decrease by 8.7\% as the number of objects in the scene increases, possibly because it is difficult for dialogue models to process long texts in larger scenes.
In contrast, the task accuracy of GRID fluctuates within $62.2\% \pm 2\%$ indicating that our network can generalize well to scenes of different sizes (with different numbers of objects).

\textbf{Generalization to unseen scenes.}
To evaluate the generalization ability of our model on unseen scenes, we separately constructed a dataset where all scenes had not appeared during model training.
The results in Tab. \ref{tab-Unseen_Scene} show that, compared to the 70-object dataset, the subtask accuracy in the unseen scenes dataset only decreases by about 2\%, and the task accuracy decreases by about 3\%.
This demonstrates that our model has relatively good generalization performance for different scenes without any additional training.
\begin{table}[h]
    \center
    \begin{tabular}{ccccc}
        \hline
        \textbf{Dataset}      & \textbf{Act. Acc.} & \textbf{Obj. Acc.} & \textbf{Sub. Acc.} & \textbf{Task Acc.} \\ \hline
        70-object      & 84.8                     & 94.1                     & 83.0                      & 64.1                   \\
        Unseen Scenes & 84.1                     & 93.7                     & 81.9                      & 61.3                  \\ \hline
    \end{tabular}
    \caption{Generalization to unseen scenes.}
    \vspace{-10pt}
    \label{tab-Unseen_Scene}
\end{table}

\subsection{Simulation Experiment}
\begin{figure}[H]
    \centering
    \includegraphics[width=0.6\columnwidth]{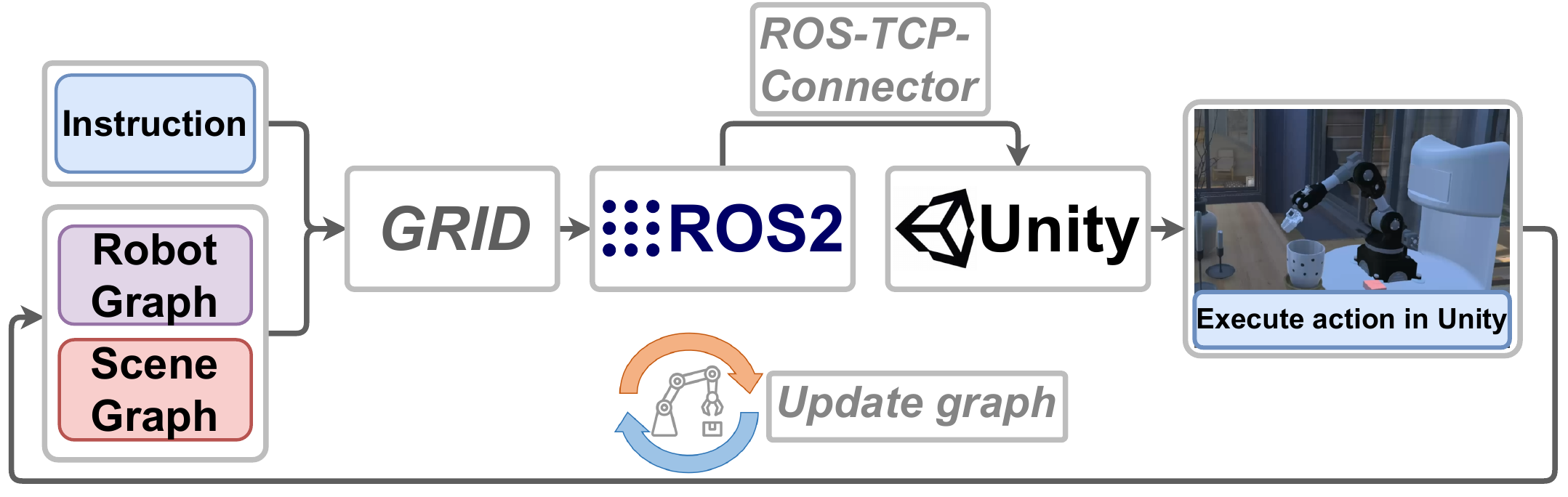}
    \caption{Experimental system design in simulation.}
    \vspace{-10pt}
    \label{fig-Simulation_Setup}
\end{figure}

\textbf{System Setup.}
 To validate the functionality of the GRID network, we developed a robot simulation platform using Unity to conduct simulation experiments. 
The simulation system design is illustrated in Fig. \ref{fig-Simulation_Setup}. 
GRID and the robot control algorithm are deployed in ROS2 and interact with the physics simulation environment in Unity through the ROS-TCP Connector communication framework.

\begin{figure}[H]
    \centering
    \begin{subfigure}{0.32\columnwidth}
        \centering
            \includegraphics[height=\linewidth]{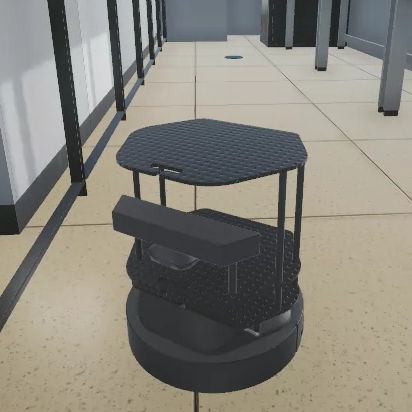}
    \end{subfigure}
    \begin{subfigure}{0.32\columnwidth}
        \centering
            \includegraphics[height=\linewidth]{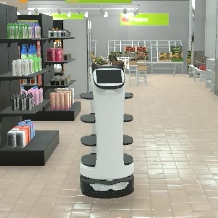}
    \end{subfigure}
    \begin{subfigure}{0.32\columnwidth}
        \centering
            \includegraphics[height=\linewidth]{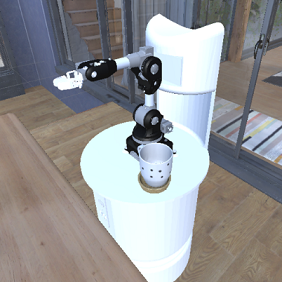}
    \end{subfigure}
        
    \caption{Simulation demonstration.}
    \vspace{-10pt}
    \label{fig-Simu_Robots}
\end{figure}

\begin{figure}[H]
    \centering
    \begin{subfigure}{0.32\columnwidth}
        \centering
        \begin{minipage}{\linewidth}
            \includegraphics[width=\linewidth]{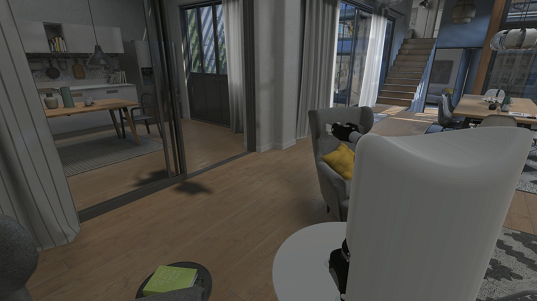}
            \subcaption{Go to dining room}
        \end{minipage}
    \end{subfigure}
    \begin{subfigure}{0.32\columnwidth}
        \centering
        \begin{minipage}{\linewidth}
            \includegraphics[width=\linewidth]{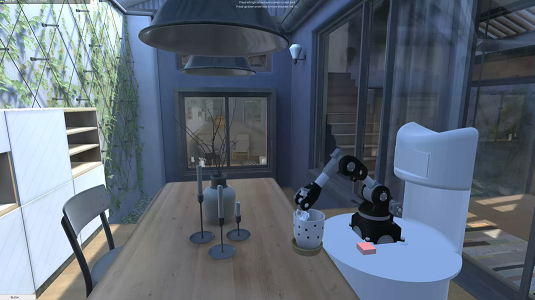}
            \subcaption{Picking up teacup}
        \end{minipage}
    \end{subfigure}
    \begin{subfigure}{0.32\columnwidth}
        \centering
        \begin{minipage}{\linewidth}
            \includegraphics[width=\linewidth]{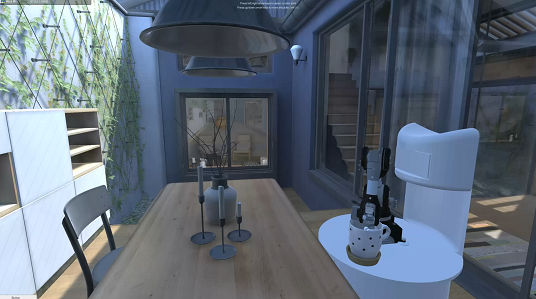}
            \subcaption{Picked up teacup}
        \end{minipage}
    \end{subfigure}

    \begin{subfigure}{0.32\columnwidth}
        \centering
        \begin{minipage}{\linewidth}
            \includegraphics[width=\linewidth]{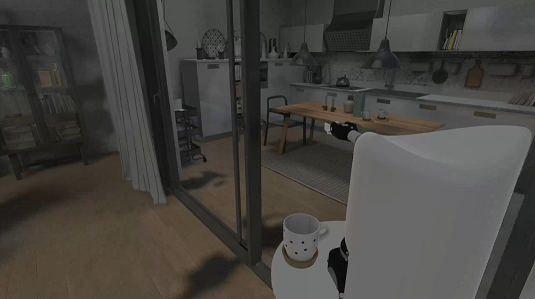}
            \subcaption{Go to kitchen}
        \end{minipage}
    \end{subfigure}
    \begin{subfigure}{0.32\columnwidth}
        \centering
        \begin{minipage}{\linewidth}
            \includegraphics[width=\linewidth]{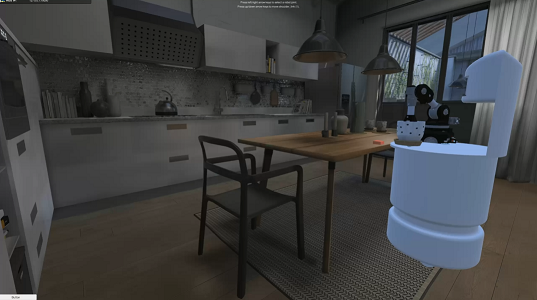}
            \subcaption{Placing teacup}
        \end{minipage}
    \end{subfigure}
    \begin{subfigure}{0.32\columnwidth}
        \centering
        \begin{minipage}{\linewidth}
            \includegraphics[width=\linewidth]{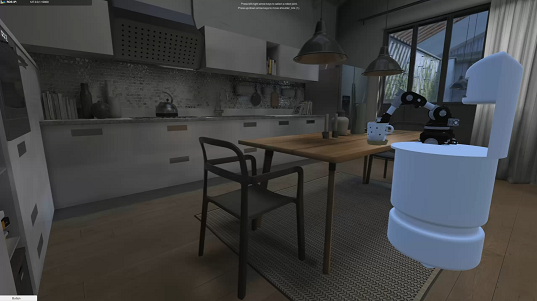}
            \subcaption{Done all subtask}
        \end{minipage}
    \end{subfigure}
        
    \caption{Mobile-pick simulation process.}
    \vspace{-10pt}
    \label{fig-Mobile_Pick_Simulation_Proce}
\end{figure}

\textbf{Simulation Demonstration.}
We deployed GRID on different types of robots (turtlebot, mobile robot, mobile-pick robot, etc.) and tested it in various scenarios (office, shop, home, etc.), as shown in Fig. \ref{fig-Simu_Robots}.
Fig. \ref{fig-Mobile_Pick_Simulation_Proce} illustrates an exemplary scenario where a mobile-pick robot successfully executes a complete task within the simulation environment. 
The provided language instruction for this example is as follows: "Please help me take the teacup from the dining table to the kitchen table." 
Subfigures (a) to (f) demonstrate the step-by-step process of the robot's movement, starting from the living room, then proceeding to the dining room to retrieve the teacup, and finally advancing to the kitchen for its placement.
This demonstration indicates that our approach can be deployed on a mobile-pick robot and perform robotic task planning based on a given human-language instruction.

\subsection{Real Deployment Experiment}
\begin{figure}[H]
    \centering
    \includegraphics[width=0.6\columnwidth]{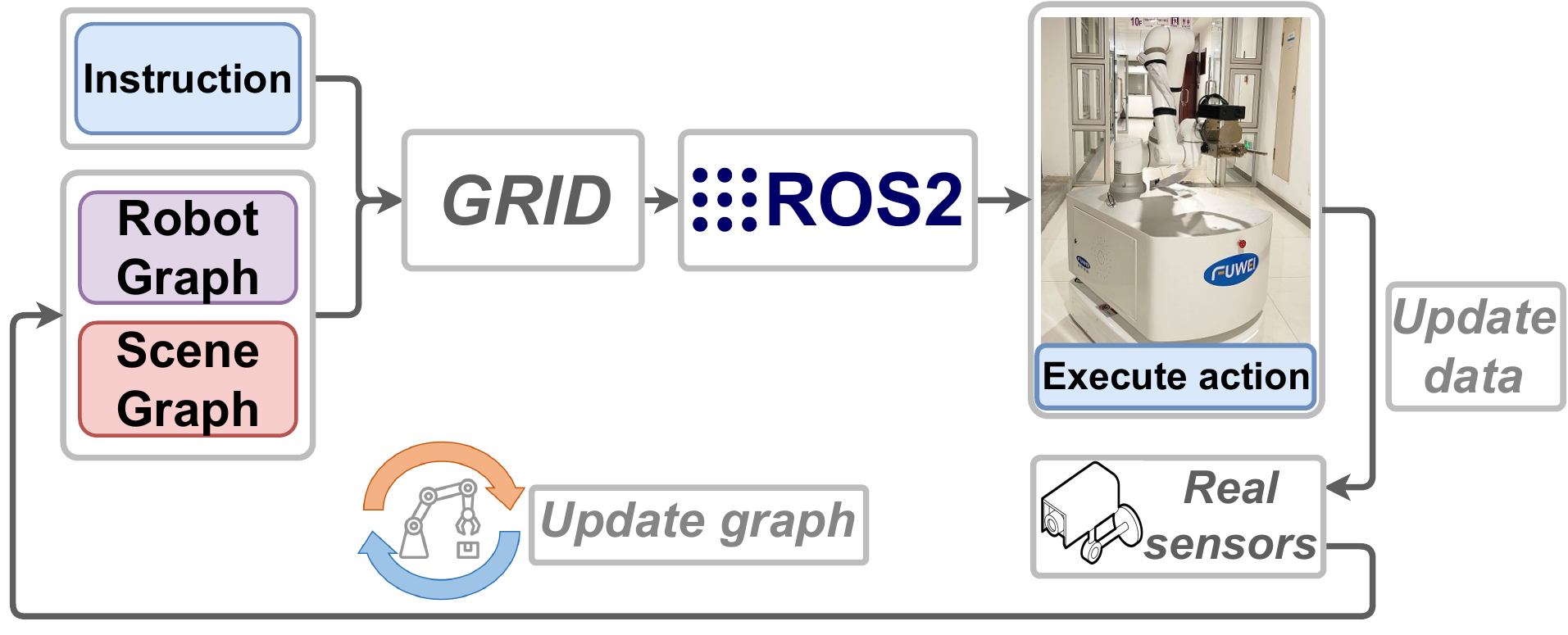}
    \caption{Experimental system design in real world.}
    \vspace{-10pt}
    \label{fig-Real_Setup}
\end{figure}

\textbf{System Setup.}
To validate the efficacy of our work in real robot motion planning and explore its potential for handling a wider range of tasks in the future, we devised a process for deploying the GRID network onto physical robots for experimental purposes. 
The scene knowledge is constructed into a scene graph using scene graph generation methods\cite{armeni_3d_2019}, wherein the pose of objects can be obtained by a 6DoF pose estimation model \cite{zhu_hierarchical_2021}.
The system structure is similar to the simulation experiment, as depicted in Fig. \ref{fig-Real_Setup}.

\begin{figure}[H]
    \centering
    \begin{subfigure}{0.29\columnwidth}
        \centering
            \includegraphics[height=\linewidth]{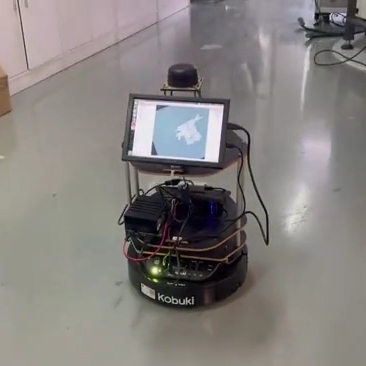}
    \end{subfigure}
    \begin{subfigure}{0.29\columnwidth}
        \centering
            \includegraphics[height=\linewidth]{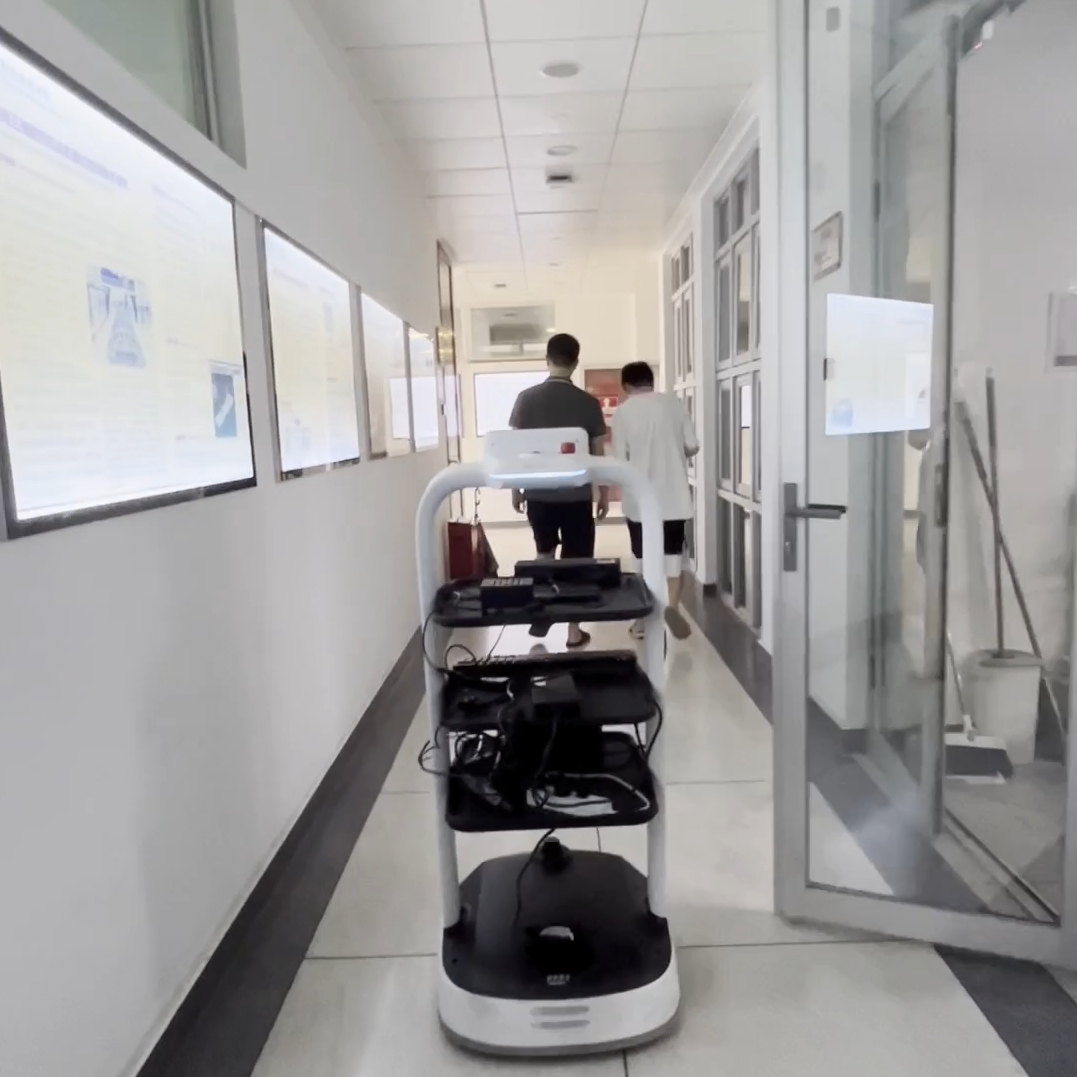}
    \end{subfigure}
    \begin{subfigure}{0.29\columnwidth}
        \centering
            \includegraphics[height=\linewidth]{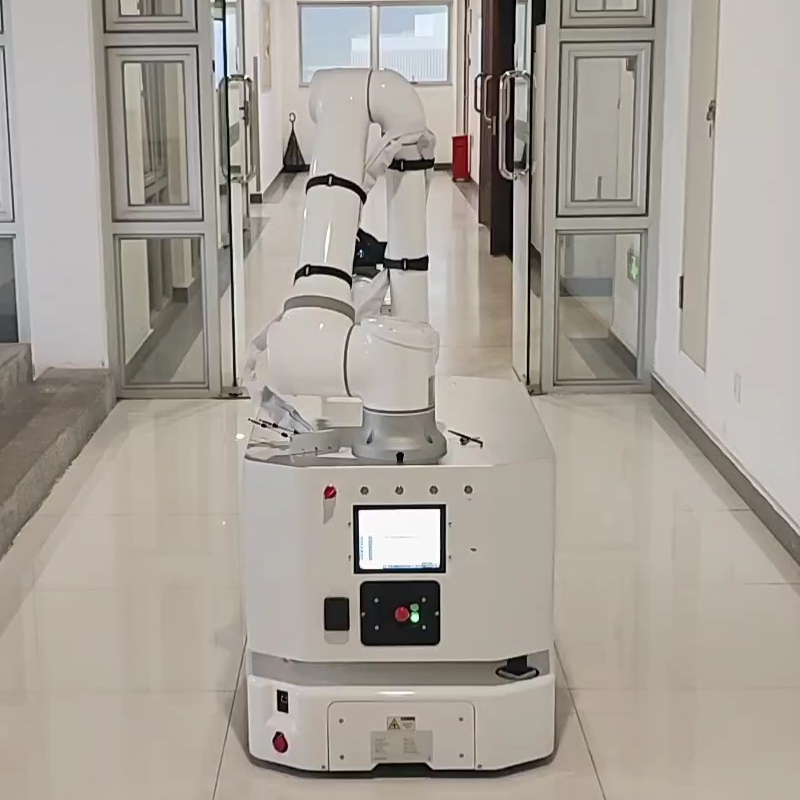}
    \end{subfigure}
        
    \caption{Real world demonstration.}
    \vspace{-10pt}
    \label{fig-Real_Robots}
\end{figure}

\begin{figure}[H]
    \centering
    \begin{subfigure}{0.29\columnwidth}
        \centering
        \begin{minipage}{\linewidth}
            \centering
            \includegraphics[width=\linewidth]{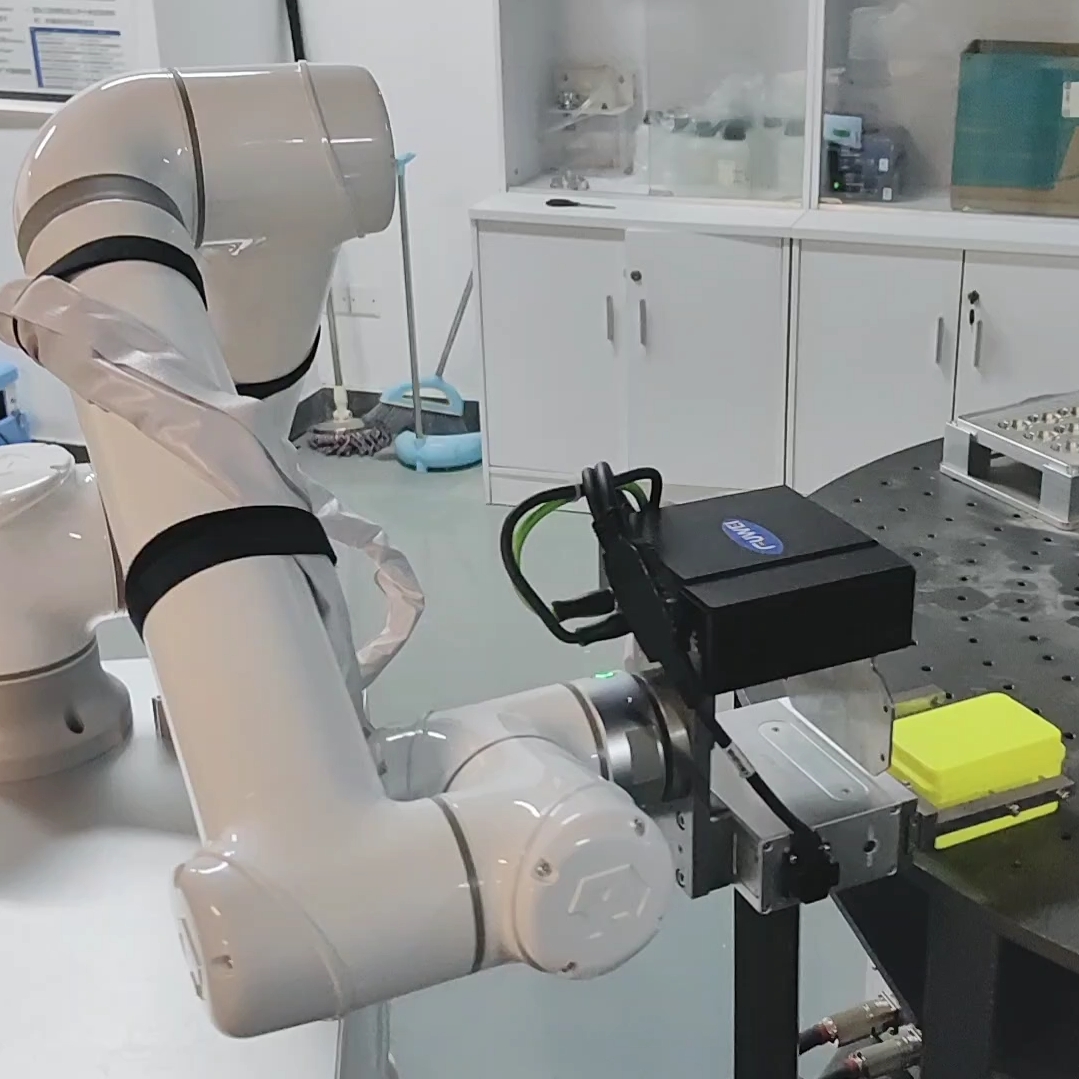}
            \subcaption{Picking up box}
        \end{minipage}
    \end{subfigure}
    \begin{subfigure}{0.29\columnwidth}
        \centering
        \begin{minipage}{\linewidth}
            \centering
            \includegraphics[width=\linewidth]{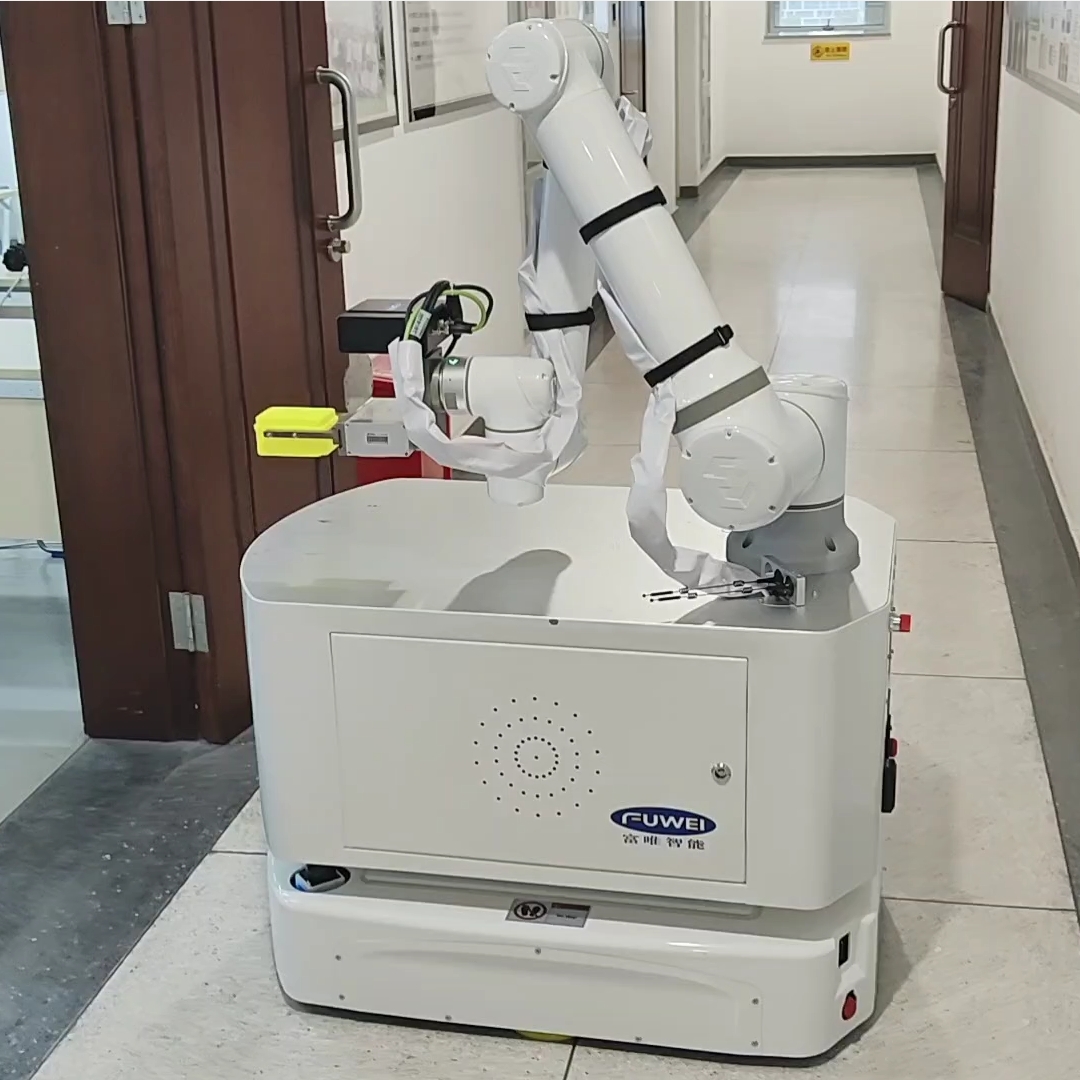}
            \subcaption{Moving}
        \end{minipage}
    \end{subfigure}
    \begin{subfigure}{0.29\columnwidth}
        \centering
        \begin{minipage}{\linewidth}
            \centering
            \includegraphics[width=\linewidth]{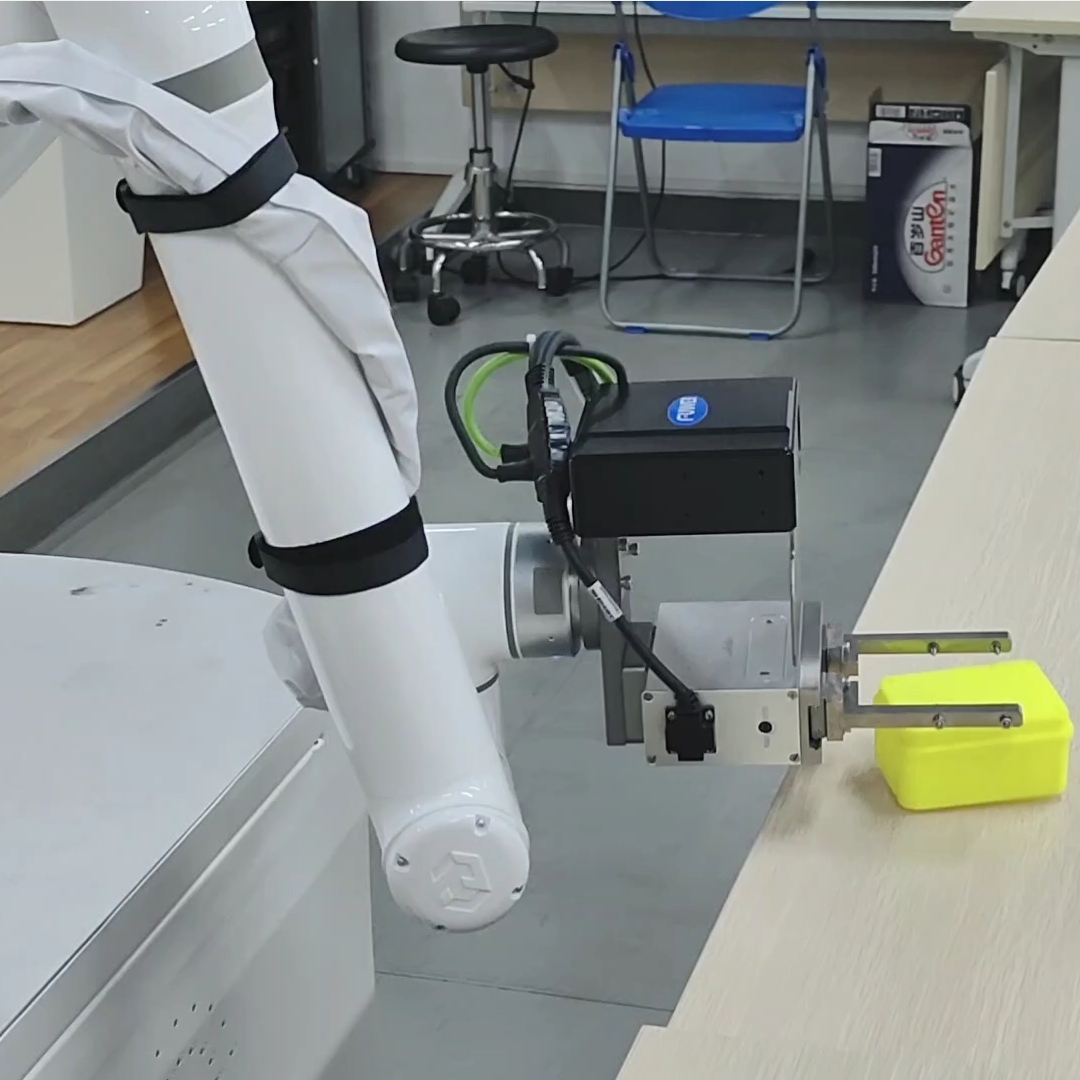}
            \subcaption{Placing box}
        \end{minipage}
    \end{subfigure}
        
    \caption{The delivery task process.}
    \vspace{-10pt}
    \label{Mobile Service Process}
\end{figure}

\textbf{Real Robot Demonstration.}
In real-world scenarios, GRID can be deployed to robots in various forms such as turtlebots, mobile robots, and mobile-pick robots, among others, as illustrated in Fig. \ref{fig-Real_Robots}. 
Fig. \ref{Mobile Service Process} demonstrates a mobile-pick robot performing a delivery task in a campus scene.
The specific language instruction: "Please bring the yellow box on the table in the laboratory to the table in the meeting room."

The above demonstration suggests that our network, GRID, empowers robots of diverse configurations to plan instruction-driven tasks and execute subtasks in real-world scenarios. 
We hope that our approach will inspire further exploration into utilizing scene graphs for instruction-driven robotic task planning. 

%% file: conclusions.tex
\section{Conclusions}

In this paper, we present a novel approach utilizing scene graphs, rather than images, as inputs to enhance the comprehension of the environment in robotic task planning. 
At each stage, our network, GRID, receives instructions, robot graphs, and scene graphs as inputs, producing subtasks in \texttt{<action>-<object>} pair format.
GRID incorporates a frozen LLM for semantic encoding and employs GAT modules to consider relationships between objects. 
The results indicate superior performance of our method over GPT-4 in scene-graph-based instruction-driven robotic task planning, showcasing adaptability to diverse scenes.